\documentclass{article}

\usepackage[preprint]{neurips_2024}

\usepackage{natbib}
\setcitestyle{numbers,square}

\usepackage[utf8]{inputenc} 
\usepackage[T1]{fontenc}    
\usepackage{hyperref}       
\usepackage{url}            
\usepackage{booktabs}       
\usepackage{amsfonts}       
\usepackage{nicefrac}       
\usepackage{microtype}      
\usepackage{xcolor}         
\usepackage{algorithmic}
\usepackage{algorithm}
\usepackage{amsmath}
\usepackage{mathabx}
\usepackage{comment}
\usepackage{graphicx}
\usepackage{enumitem}
\newcommand{\DualUpdate}{Disocclusion Boundary Re-Injection}
\newcommand{\dualupdate}{disocclusion boundary re-injection}
\newcommand{\Vid}{\mathbf{X}}
\newcommand{\Frame}{X}
\newcommand{\VidDepth}{\mathbf{d}}
\newcommand{\FrameSub}{s}
\newcommand{\FrameSubMax}{S}
\newcommand{\Prompt}{c}
\newcommand{\LeftSub}{l}
\newcommand{\RightSub}{r}

\newcommand{\Generator}{\mathcal{G}}
\newcommand{\Noise}{\mathbf{\epsilon}}
\newcommand{\VidLatent}{\mathbf{z}}
\newcommand{\ImageMasks}{\mathbf{M}}

\newcommand{\LatentMasks}{\mathbf{m}}
\newcommand{\LatentMask}{m}
\newcommand{\ViewSub}{v}
\newcommand{\ViewSubMax}{V}
\newcommand{\NoiseScale}{\bar{\alpha}_t}
\newcommand{\Encoder}{\mathcal{E}}
\newcommand{\Decoder}{\mathcal{D}}
\newcommand*{\vertbar}{\rule[-1ex]{0.5pt}{2.5ex}}
\newcommand*{\horzbar}{\rule[.5ex]{2.5ex}{0.5pt}}

\title{SVG: 3D Stereoscopic Video Generation via Denoising Frame Matrix}

\author{%
  Peng Dai$^{1,2}$\quad \textbf{Feitong Tan}$^{1}$\thanks{Equal contribution}\quad
  \textbf{Qiangeng Xu}$^{1*}$\quad \textbf{David Futschik}$^{1}$\\
  \textbf{Ruofei Du}$^{1}$\quad \textbf{Sean Fanello}$^{1}$\quad \textbf{Xiaojuan Qi}$^{2}$\quad \textbf{Yinda Zhang}$^{1}$\\
  $^{1}$Google\quad $^{2}$The University of Hong Kong}

\begin{document}

\maketitle

\begin{abstract}
Video generation models have demonstrated great capabilities of producing impressive monocular videos, however, the generation of 3D stereoscopic video remains under-explored. We propose a pose-free and training-free approach for generating 3D stereoscopic videos using an off-the-shelf monocular video generation model. Our method warps a generated monocular video into camera views on stereoscopic baseline using estimated video depth, and employs a novel \textit{frame matrix} video inpainting framework. The framework leverages the video generation model to inpaint frames observed from different timestamps and views. This effective approach generates consistent and semantically coherent stereoscopic videos without scene optimization or model fine-tuning. Moreover, we develop a \dualupdate~scheme that further improves the quality of video inpainting by alleviating the negative effects propagated from disoccluded areas in the latent space. We validate the efficacy of our proposed method by conducting experiments on videos from various generative models, including Sora~\cite{videoworldsimulators2024}, Lumiere~\cite{bar2024lumiere}, WALT~\cite{gupta2023photorealistic}, and Zeroscope~\cite{wang2023modelscope}. The experiments demonstrate that our method has a significant improvement over previous methods. The code will be released at \url{https://daipengwa.github.io/SVG_ProjectPage/}

\end{abstract}

\section{Introduction}
As VR/AR technology advances, the demand for creating stereoscopic content and delivering immersive 3D experiences to users continues to grow. 
Due to visual sensitivity, binocular stereoscopic content should feature flawless 3D and semantic consistency between both eye views, as well as seamless temporal consistency across frames. 
While monocular video generation models have been extensively researched and methods are now capable of synthesizing high-fidelity videos that adhere to complex text prompts~\cite{videoworldsimulators2024}, there has not been much progress in the realm of generating 3D stereoscopic videos at the scene level.
One reason for this gap lies in the substantial amount of monocular video data that is readily available, contrasted with the scarcity of stereo video data for training models to generate stereoscopic videos directly.

An emergent solution is to convert generated monocular videos into stereoscopic videos using novel view synthesis~\cite{li2023dynibar, liu2023robust}.
However, these methods usually overly rely on camera pose estimation, which is a challenging task on its own either using SFM~\cite{schoenberger2016sfm} or joint optimization~\cite{liu2023robust}, and as a result tend to be unstable, particularly in dynamic scenes where cameras experience subtle motions or when the content is dominated by dynamic objects with temporally varying appearances, both of which are prevalent in generated videos.
Consequently, these methods fail in optimizing 3D scenes and offer low-quality solutions to the task (see Fig.~\ref{fig:comparisons}).
Moreover, these approaches are based on reconstruction, lacking the generative ability to hallucinate occluded regions in the novel views that do not appear in any of the remaining video frames. 



In this paper, we propose an alternative pose-free and training-free framework, for the sake of robustness and generalization capability,
that operates solely by exploiting inference of an off-the-shelf video generation model~\cite{wang2023modelscope} to generate high quality 3D stereoscopic videos. 
Our initial attempt follows a typical 2D to 3D image uplifting methodology~\cite{hollein2023text2room} and extends it into the video domain.
Specifically, we first generate a monocular video as the left view,
which is then reprojected into the right view using per-frame estimated monocular depths~\cite{yang2024depth}, where we apply temporal-spatial smoothing to improve the consistency of the estimated depth.
Subsequently, we leverage an off-the-shelf video generation model's~\cite{wang2023modelscope} ability to generate natural videos, by adding noise and denoising the warped video frames to inpaint the disoccluded regions, inspired by diffusion-based image inpainting~\cite{avrahami2023blended}. 

However, this naive pipeline does not produce appealing results:
inpainting the right-view video frames independently, without referencing the left view, typically generates semantically mismatched content.
To address this problem, we propose a novel representation, called the \textit{frame matrix}, which contains frame sequences observed from a number of viewpoints evenly distributed along the baseline between two eyes. The frame sequences along the view direction  (rows of the matrix) form videos with camera motion, while the frame sequences along the time direction (columns of the matrix) form videos with scene motions (see Fig.~\ref{fig:pipeline} second column). Since the video diffusion model has video prior for both scene and camera motions, we propose to jointly update the entire \textit{frame matrix} from both directions. In each denoising step, we use resample techniques~\cite{lugmayr2022repaint} by alternatively denoising frame sequences along the view and the time directions. Finally, we obtain a semantically consistent and temporally smooth 3D stereoscopic video by taking the leftmost and the rightmost frame sequences to represent the left-eye view and the right-eye view, respectively.
Furthermore, we note that the inevitable resolution downsampling operation in most video generation models with latent encoding~\cite{videoworldsimulators2024, bar2024lumiere, wang2023modelscope, gupta2023photorealistic} is detrimental to the video inpainting task. During encoding, the dark pixels created by disocclusion can degrade the features near the disocclusion boundary, leading to undesirable artifacts  (see Fig.~\ref{fig:update_feature}). Instead of following the inpainting scheme proposed in previous work~\cite{avrahami2023blended}, which encodes the latent feature only once, we iteratively update both the disoccluded regions in the image space and the latent feature map with generated content during the diffusion process. This approach re-injects the generated content into the disocclusion boundary, which mitigates the negative impact of dark disocclusion and effectively prevents the artifacts.

%


To validate the efficacy of our proposed method, we generate stereoscopic video from monocular videos generated by Sora~\cite{videoworldsimulators2024}, Lumiere~\cite{bar2024lumiere}, WALT~\cite{gupta2023photorealistic}, and Zeroscope~\cite{wang2023modelscope}. Both qualitative and quantitative evaluations suggest that our approach outperforms other baselines in 3D stereoscopic video generation. Our contributions are summarized as follows:
\begin{itemize}[topsep=0pt]
    \item We design a novel pipeline to generate 3D stereoscopic videos. Unlike previous work, our method does not need camera pose estimation or fine-tuning on specific datasets. 
    \item We propose a novel \textit{frame matrix} representation that regularizes the diffusion-based video inpainting to generate semantically consistent and temporally smooth content. 
    \item We propose a re-injection scheme that drastically reduces the negative influence of disoccluded regions in latent space and produces high-quality results. 
    \item We conduct comprehensive experiments that show the superiority of our approach over previous methods for 3D stereoscopic video generation. 
\end{itemize}

\section{Related Work}
\textbf{Video Generation.}
Video generation~\cite{wang2023modelscope, videoworldsimulators2024, bar2024lumiere, gupta2023photorealistic, harvey2022flexible, ho2022imagen, ho2022video, singer2022make} has achieved tremendous progress since the advent of the diffusion model~\cite{ho2020denoising}. Taking into account the dataset requirements and scarcity of tagged videos, a prominent approach for video generation is to extend pre-trained image generation models~\cite{rombach2022high, saharia2022photorealistic, ramesh2022hierarchical} by inserting additional temporal layers and then fine-tuning them on video data~\cite{guo2023animatediff, blattmann2023align, wu2023tune}. To further improve the compute efficiency and enable long clip processing, WALT~\cite{gupta2023photorealistic} and Lumiere~\cite{bar2024lumiere} proposed to compress the video in both the temporal and spatial dimensions. 
More recently, Sora~\cite{videoworldsimulators2024} adopted a transformer diffusion architecture~\cite{peebles2023scalable} and was trained on large-scale video datasets to produce impressive video generation results.
Different from previous video generation models focusing on producing higher-quality and longer monocular videos, our method orthogonally explores the possibility of leveraging pre-trained video generation models for stereoscopic 3D video generation.   

\textbf{Novel View Synthesis.} Great progress has been made for novel view synthesis in both static and dynamic scenes captured by single or multiple cameras~\cite{mildenhall2021nerf, yoon2020novel, li2022neural, kerbl20233d, muller2022instant}. Mildenhall \emph{et al.}~\cite{mildenhall2021nerf} proposed to encode the static scene into neural radiance fields (NeRF), which were then used for novel view synthesis through volume rendering. For more challenging scenes with dynamic content, follow-up works additionally optimized a deformation field~\cite{park2021nerfies, huang2023sc, park2021hypernerf} or scene flow fields~\cite{li2021neural} to handle the motion of dynamic objects. Instead of encoding the scene into a NeRF, DynIBaR~\cite{li2023dynibar} leveraged nearby frames for rendering novel view images, and dynamic objects were handled by optimized motion fields. Different from methods requiring pre-computed camera poses, RoDynRF~\cite{liu2023robust} jointly optimized the NeRF and camera poses from scratch. Concurrently, FVS~\cite{lee2023fast} achieves novel view video synthesis using a plane-based scene representation. Although these approaches produce high-quality renderings, they are limited to scenes where the camera pose can be accurately estimated and have limited synthesis capability. In contrast, we design a method that explicitly avoids having to estimate camera poses and possesses the ability to hallucinate unseen content.

\textbf{3D Content Creation and Inpainting.} Automated 3D content creation~\cite{hollein2023text2room, dai2020neural, gao2024gaussianflow, yu2023wonderjourney} is another related area, with emerging approaches such as inpainting~\cite{ho2022imagen} or multi-view generators~\cite{liu2023zero, wang2024stereodiffusion}. Recently, Text2Room~\cite{hollein2023text2room} proposed creating a 3D room by warping an image into novel views and using a text-guided inpainter to deal with disocclusions. WonderJourney~\cite{yu2023wonderjourney} made this process automatic by including a large language model in the loop. Similar to creating static scenes, we could use pretrained video inpainter~\cite{zhou2023propainter, li2022towards} for dynamic 3D content creation, however, these models suffer from generalization problems in creating high-quality, consistent 3D content. Lastly, Deep3D~\cite{xie2016deep3d} is trained using 3D movies, with the goal of converting 2D videos into stereoscopic videos. However, the training data is not publicly available and it lacks the flexibility to modify videos for creative purposes, such as different stereo baselines. In this paper, we explore the possibilities of using video generation models for 3D video creation without training on specific, hard-to-obtain datasets.

\section{Stereoscopic Video Generation}
Conditioned on a text prompt or a single image $\Prompt$, our method aims to generate 3D stereoscopic video $\{\Vid_{\LeftSub},\Vid_{\RightSub}\}$, consisting of two monocular sequences.
The most straightforward way is to use a diffusion-based generation model $\Generator$:
\begin{align}
   \{\Vid_{\LeftSub},\Vid_{\RightSub}\} = \Generator(\{\Noise_t | t=1,..., T\}, c), 
    \label{eq:ours_3}
\end{align}
where $\Noise_t \sim \mathcal{N}(\mathbf{0}, \mathbf{I})$ is the sampled noise at step $t$.
The generated stereoscopic videos should possess the following characteristics: First, the appearance and semantics between the left eye view $\Vid_{\LeftSub}$ and right eye view $\Vid_{\RightSub}$ should be consistent and be temporally stable. Second, the stereo effect should be prominent and immersive. Last, the generated content should be diverse and controllable with the given conditioning. 

However, training a $\Generator$ that can directly generate stereo videos $\{\Vid_{\LeftSub},\Vid_{\RightSub}\}$ with the desired properties requires a vast dataset of stereo videos with diverse content. Due to the scarcity of such data, we propose a training-free approach that relies on an off-the-shelf depth estimator \cite{yang2024depth} and a diffusion-based monocular video generation model $\Generator$ such as Zeroscope~\cite{wang2023modelscope}. We first generate a monocular video for one eye using a video diffusion model \cite{wang2023modelscope, gupta2023photorealistic, videoworldsimulators2024, bar2024lumiere} (Eq. \ref{eq:leftgen}), then obtain the other video view by conditioning on the first video. To automatically preserve 3D consistency, we implement this conditioning by estimating depth $\VidDepth_\LeftSub$ for the left video and warp its content to obtain the right view sequence $\Vid_{\LeftSub\rightarrow\RightSub}$ with disocclusion masks $\ImageMasks_\RightSub$ (Eq. \ref{eq:warp}) according to the stereoscopic baseline. Then, we use $\Generator$ again to inpaint the disoccluded parts by denoising inpainting process \cite{avrahami2023blended,lugmayr2022repaint} (Eq. \ref{eq:inpaint}), obtaining the other eye view video $\Vid_{\RightSub}$.
\begin{align}
    \Vid_{\LeftSub} &= \Generator(\{\Noise_t | t=1,..., T\}, c), \label{eq:leftgen} \\
    \Vid_{\LeftSub\rightarrow\RightSub}, \ImageMasks_\RightSub &= \textrm{Warp}_{\LeftSub\rightarrow\RightSub}(\Vid_{\LeftSub}, \VidDepth_\LeftSub),
    \label{eq:warp} \\
    \Vid_{\RightSub} &= \Generator(\{\Noise_t | t=1,..., T\}, c, \Vid_{\LeftSub\rightarrow\RightSub}, \ImageMasks_\RightSub).
    \label{eq:inpaint}
\end{align}
\vspace{-20pt}

\begin{figure}
    \centering
    \includegraphics[width=1.0\linewidth]{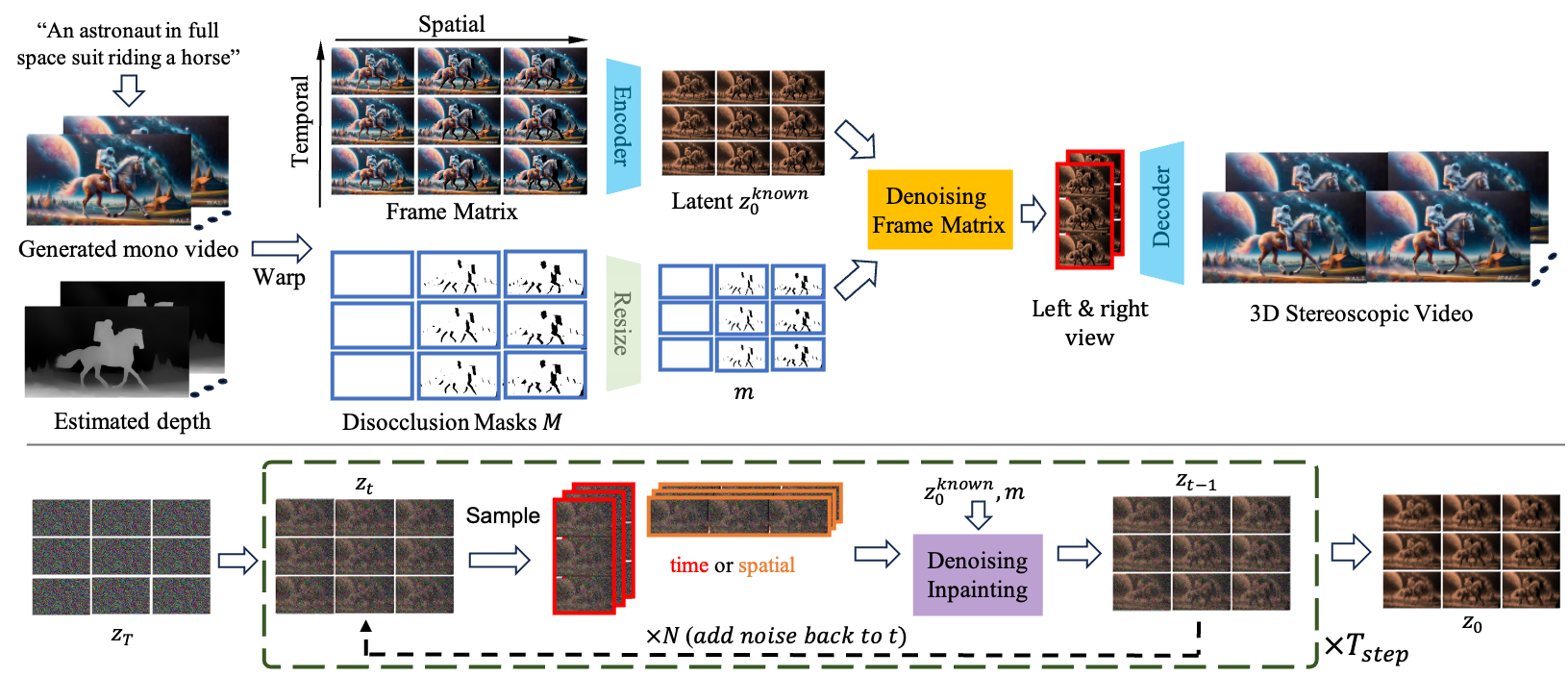}
    \vspace{-15pt}
    \caption{\textbf{Overview} -- \textbf{Top:} Given a text prompt, our method first uses a video generation model to generate a monocular video, which is warped (using estimated depth) into pre-defined camera views to form a \textit{frame matrix} with disocclusion masks $M$. Then, the disoccluded regions are inpainted by denoising the frame sequences within the \textit{frame matrix}. After denoising, we select the leftmost and the rightmost columns and decode them to obtain a 3D stereoscopic video. \textbf{Bottom:} Details of denoising \textit{frame matrix}. We initialize the latent matrix $\VidLatent_T$ as a random noise map. For each noise level, we extend the resampling mechanism  \cite{karnewar2023holodiffusion,lugmayr2022repaint} to alternatively denoise temporal (column) sequences and spatial (row) sequences $N$ times. Each time, row or column sequences are denoised and inpainted (see Fig.\ref{fig:denoising_inpainting}). By denoising along both spatial and temporal directions, we obtain an inpainted latent $\VidLatent_0$ which can be decoded into temporally smooth and semantically consistent sequences.}
    \vspace{-10pt}
    \label{fig:pipeline}
\end{figure}

In Sec.~\ref{sec:video_warp}, we describe the video depth warping. In Sec.~\ref{sec:frame_matrix}, we introduce the \textit{frame matrix} representation for the video inpainting. Our denoising frame matrix drastically improves the semantic similarity between $\Vid_{\LeftSub}$ and $\Vid_{\RightSub}$ and helps preserve temporal smoothness. Last but not least, a \dualupdate~mechanism is introduced to further improve the inpainting quality in Sec.~\ref{sec:boundary_reinjection}. An overview of our method is displayed in Fig.~\ref{fig:pipeline}.


\subsection{Monocular Video Depth Warping}
\label{sec:video_warp}
The depth estimation model~\cite{yang2024depth} is applied to predict all frames' depth values, which will be smoothed to produce more consistent video depths. Specifically, we utilize the estimated optic flows~\cite{teed2020raft} to align consecutive depth frames. The outliers in predicted depths will be suppressed by convolving with a Gaussian kernel along the time axis. 
After obtaining RGB-D frames, we can warp them into target camera views where disoccluded regions appear.
In addition, the warped images usually contain isolated pixels, and the foreground and background are entangled, which jeopardizes video quality~\cite{dai2020neural}. To handle these problems, we follow Dai~\emph{et al.}~\cite{dai2020neural} to project points into multi-plane images~\cite{zhou2018stereo}, then remove isolated pixels and cracks and finally obtain a noisy-points-free image. (See supplemental material for details).        

\vspace{-5pt}

\subsection{Video Inpainting with Frame Matrix}
\label{sec:frame_matrix}
\begin{figure}
    \centering
    \includegraphics[width=1.0\linewidth]{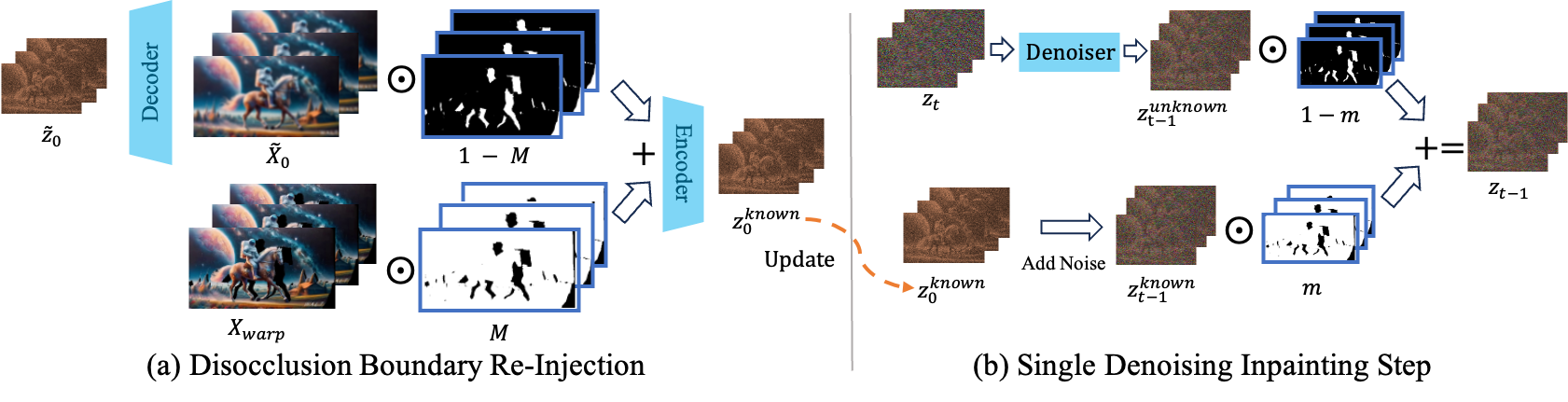}
    \vspace{-15pt}
    \caption{\textbf{Denosing Inpainting}. This figure visualizes the operations in the purple box of Fig.\ref{fig:pipeline}. (a) We re-inject the generated content from a denoised latent $\widetilde{\VidLatent}_0$ to update $\VidLatent_0^{known}$ and reduce its feature corruption on the disocclusion boundary. (b) A noisy latent $\VidLatent_t$ is denoised to $\VidLatent_{t-1}^{unknown}$. We take its disoccluded region and combine it with the unoccluded region of $\VidLatent_0^{known}$. }
    \vspace{-10pt}
    \label{fig:denoising_inpainting}
\end{figure}

The inpainting pipeline plays a key role in ensuring spatial/semantic and temporal consistency. While image inpainting approaches~\cite{avrahami2023blended, lugmayr2022repaint} provide a reasonable baseline, the results lack temporal and spatial stability. Therefore, we introduce a Frame Matrix representation, which addresses both issues.
\paragraph{Single Video Denoising Inpainting} 
Inspired by RePaint~\cite{lugmayr2022repaint}, we extend the diffusion-based image inpainting to video inpainting. We use the video generation model $\Generator$ (i.e., Zeroscope~\cite{wang2023modelscope}) as our inpainting tool, which is a latent diffusion model consisting of a VAE encoder $\Encoder$, a decoder $\Decoder$ and a latent denoiser $\{ \epsilon_{\theta}, \Sigma_{\theta}\}$. First, the warped video is fed into the VAE encoder to obtain video latent features $\VidLatent_0^{\text{known}} = \Encoder(\Vid_{\LeftSub\rightarrow\RightSub})$. 
Then, we resize the image disocclusion masks $\ImageMasks_\RightSub$ to the resolution of the latent and obtain latent disocclusion masks $\LatentMasks$. During the denoising process, we start from a random noisy latent map $\VidLatent_T \sim~ \mathcal{N}(\mathbf{0}, \mathbf{I})$. For each subsequent step $t$, we sample a new intermediate noisy latent map from $\VidLatent_0$ (Eq. \ref{eq:inpaintknown}), denoises the latent map from the last step $\VidLatent_t$ (Eq. \ref{eq:inpaintunknown}) and combine them with $\LatentMasks$ to obtain the $\VidLatent_{t-1}$ (Eq. \ref{eq:inpaintt_1}). We visualize the following steps in Fig.\ref{fig:denoising_inpainting} (b):
\begin{align}
    \VidLatent_{t-1}^{\text{known}} \sim&~ \mathcal{N}\left(\sqrt{\NoiseScale}\VidLatent_0^{\text{known}}, (1 - \NoiseScale)\mathbf{I}\right),
    \label{eq:inpaintknown} \\
     \VidLatent_{t-1}^{\text{unknown}} \sim&~ \mathcal{N}\left(\frac{1}{\sqrt{1-\beta_t}}\left(\VidLatent_t - \frac{\beta_t}{\sqrt{1-\Bar{\alpha}_t}}\epsilon_{\theta}(\VidLatent_t,\Prompt,t)\right), \Sigma_{\theta}(\VidLatent_t, \Prompt, t)\right),
    \label{eq:inpaintunknown} \\
    \VidLatent_{t-1} =& ~\LatentMask ~\odot~ \VidLatent_{t-1}^{\text{known}} + (1-\LatentMask) ~\odot~ \VidLatent_{t-1}^{\text{unknown}},
    \label{eq:inpaintt_1}
\end{align}
where $\NoiseScale$ and $\beta_t$ denote the total noise variance and one step noise variance at $t$, respectively; $\epsilon_{\theta}(\VidLatent_t,\Prompt,t)$ and $\Sigma_{\theta}(\VidLatent_t,\Prompt,t)$ are predicted noise and variance for noisy latent map at $t-1$ step. Finally, we can obtain the inpainted right view sequence $\Frame_\RightSub$ by decoding the denoised latent $\Frame_\RightSub =  \Decoder(\VidLatent_0)$.

By applying the above video inpainting scheme for the right view, we implement Eq. \ref{eq:inpaint} and successfully hallucinate the disoccluded (unknown) regions while preserving the unoccluded (known) regions. The video diffusion model also ensures temporal smoothness. However, the inpainted content on the right view usually lacks semantic consistency w.r.t. the left view, as shown in the third column of Fig.~\ref{fig:sematic_matching}. This is because we only condition on the left view by depth warping, while dropping the conditioning during inpainting. 


\paragraph{Frame Matrix Representation.} 
We propose a novel representation--\emph{frame matrix}, which targets consistent dynamic content generation across space and time. As shown in Fig.~\ref{fig:pipeline} top, it is a matrix consisting of multiple frames, where each row represents frames observed from different camera poses at the same time stamp, and each column is a video recorded by a fixed camera at different time stamps. Consequently, the frame matrix can be defined as: 
\vspace{-5pt}
\[\tiny{
\Vid \equiv 
\left[
  \begin{array}{ccc}
    \vertbar &         & \vertbar \\
    \Vid_{(:,0)}    &  \ldots & \Vid_{(:,\ViewSubMax)}    \\
    \vertbar &         & \vertbar 
  \end{array}
\right]
 \equiv
\left[
  \begin{array}{ccc}
    \horzbar & {\Vid_{(0,:)}} & \horzbar \\
             & \vdots    &          \\
    \horzbar & \Vid_{(\FrameSubMax,:)} & \horzbar
  \end{array}
\right]}
\]
\vspace{-10pt}

where $\FrameSubMax$ and $\ViewSubMax$ are the largest indices of time steps and views, respectively. A view sequence (row) $\Vid_{(\FrameSub,:)}$ forms a video with camera motions, while a time sequence (column) $\Vid_{(:,\ViewSub)}$ forms a video with time-varying scene motions. Since the video diffusion model can denoise a sequence to a temporally and semantically consistent video, jointly denoise the rows and columns can ensure consistency spatially and temporally. Finally, we can obtain a 3D stereoscopic video by taking the leftmost and the rightmost time sequences ${\Vid_{(:,0)}, \Vid_{(:,\ViewSubMax)}}$.




\paragraph{Constructing Frame Matrix.} We evenly add $\ViewSubMax$ camera views distributed along the baseline between the two eyes with the same orientation of the reference view. Then, we warp the refence video (the $0th$ column) based on depth (Sec.~\ref{sec:video_warp}) into these views and obtain $\Vid_{warp} \equiv [\Vid_{(:,0)}, \Vid_{(:,0 \rightarrow 1)}, ..., \Vid_{(:,0 \rightarrow \ViewSubMax)}]$ with a disocclusion masks matrix $\ImageMasks$.

\vspace{-0pt}

\paragraph{Denoising Frame Matrix.}
Similar to single video sequence inpainting, we encode frame matrix into a latent frame matrix $\VidLatent_0^{\text{known}} = \Encoder(\Vid_{warp})$, and resize $\ImageMasks$ to obtain latent disocclusion map $\LatentMasks$. We also initialize $\VidLatent_T \sim~ \mathcal{N}(\mathbf{0}, \mathbf{I})$. As shown in Fig.\ref{fig:pipeline} (Bottom), for each noise level, we extend the resampling mechanism \cite{lugmayr2022repaint} to alternatively denoise column sequences and row sequences $N$ times. Each time, row or column sequences are denoised following Eq. \ref{eq:inpaintknown}-\ref{eq:inpaintt_1} and we add back noise between every resampling iteration:
\begin{align}
    \VidLatent_{t} \sim \mathcal{N}(\sqrt{1-\beta_{t-1}}\VidLatent_{t-1}, \beta_{t-1}\mathbf{I}).
    \label{eq:add_resampling_noise}
\end{align}
Please refer to Sec.\ref{sec:pseudocode} in the supplemental material.
By denoising along these two directions alternatively, the spatial and temporal sequences will proceed toward a harmonic state in the end.

\subsection{\DualUpdate}
\label{sec:boundary_reinjection}
\vspace{-5pt}
Since most video generation models are using latent diffusion, the disoccluded dark regions of $\Vid_{warp}$ will be propagated beyond the latent mask $\LatentMasks$ during VAE encoding (\textit{e.g.}, Zeroscope downsamples by $8\times$), leading to defective latent features on $\VidLatent_0^{\text{known}}$'s disocclusion boundary. This would lead to artifacts in the final results (Fig.~\ref{fig:update_feature} left). 

We propose to re-inject the denoised information in the disoccluded regions to improve the latents on this boundary. Specifically, we predict the denoised latent features \cite{ho2020denoising}, which are decoded into a denoised video (Eq. \ref{eq:denoised0}). Then, we replace its unoccluded regions with warped pixels to form a video that is faithful to the reference view but with better disocclusion pixels. By encoding this video, we can get a updated $\VidLatent_0^{\text{known}}$ (Eq. \ref{eq:update_latent}) which alleviates corruption on the boundary:
\vspace{-5pt}
\begin{align}
    \widetilde{\Vid}_0 &= \Decoder(\widetilde{\VidLatent}_0), \text{where}~ \widetilde{\VidLatent}_0 = \frac{1}{\sqrt{\Bar{\alpha}_t}}\left(\VidLatent_t - \sqrt{1-\Bar{\alpha}_t}\epsilon_{\theta}(\VidLatent_t, \Prompt,t)\right)
    \label{eq:denoised0}, \\
    \VidLatent_{0}^{\text{known}} &= \Encoder\left(\ImageMasks ~\odot~ \Vid_{warp} + (1-\ImageMasks) ~\odot~ \widetilde{\Vid}_0\right).
    \label{eq:update_latent}
\end{align}
\vspace{-5pt}
After this, this improved $\VidLatent_0^{\text{known}}$ can be used in Eq. \ref{eq:inpaintknown} for the next iteration.


\section{Experiments}

\paragraph{Datasets.}
To validate the effectiveness of our method, we conduct experiments using a variety of recent video generation models, including Sora~\cite{videoworldsimulators2024}, Lumiere~\cite{bar2024lumiere}, WALT~\cite{gupta2023photorealistic}, and Zeroscope~\cite{wang2023modelscope}. These models produce diverse left videos from a wide range of input text prompts, covering subjects such as humans, animals, buildings, and imaginary content.
\vspace{-6pt}
\paragraph{Implementation Details.}
To ensure the stereo effect appears realistic, we normalize the up-to-scale depth values predicted by the depth estimation model \cite{yang2024depth} to a range of (1, 10) and set the baseline between left and right views to 0.08. The frame matrix is constructed by evenly placing 8 cameras between the left and right views, with each camera corresponding to a warped video sequence. Due to the limitations of the zeroscope model, we currently 
conduct experiments on video sequences with 16 frames. Following the approach of RePaint \cite{lugmayr2022repaint}, we employ DDPM \cite{ho2020denoising} as our denoising scheduler with 1000 total time steps $T$ and 50 denoising steps, resulting in 20 time step jumps per denoising step. During the initial 25 denoising steps (50 to 25), we resample 8 times at each step to establish a reasonable structure in disoccluded regions. For the remaining steps, we reduce resampling to 4 times and denoise only the right view for improved efficiency while generating stereoscopic videos. We run experiments on one A6000 GPU.

\vspace{-6pt}
\paragraph{Baselines.}
We compare our method with two families of approaches: video inpainting, and novel view synthesis from a monocular video.
For video inpainting approaches, we generate the right view in the same manner as our method using depth-guided warping. We then apply state-of-the-art methods ProPainter~\cite{zhou2023propainter} and E2FGVI~\cite{li2022towards} to inpaint the right views.
For novel view synthesis methods, we compare our results with RoDynRF~\cite{liu2023robust} and DynIBaR~\cite{li2023dynibar}, which optimize scene representations relying on camera poses. To ensure a fair comparison, given the differing 3D scales between their reconstructed scenes and our estimated depth, we select the baseline for rendering the right view by matching the median disparity of foreground regions in the resulting disparity map to that of our methods. 
We are also aware of approaches trained on dedicated datasets that directly produce the right-view given the left-view like Deep3D~\cite{xie2016deep3d}.
However it does not generalize well to the generated video, especially those in non-realistic styles, and the comparison could be found in supplemental material.


\begin{figure}
    \centering
    \includegraphics[width=1.0\linewidth]{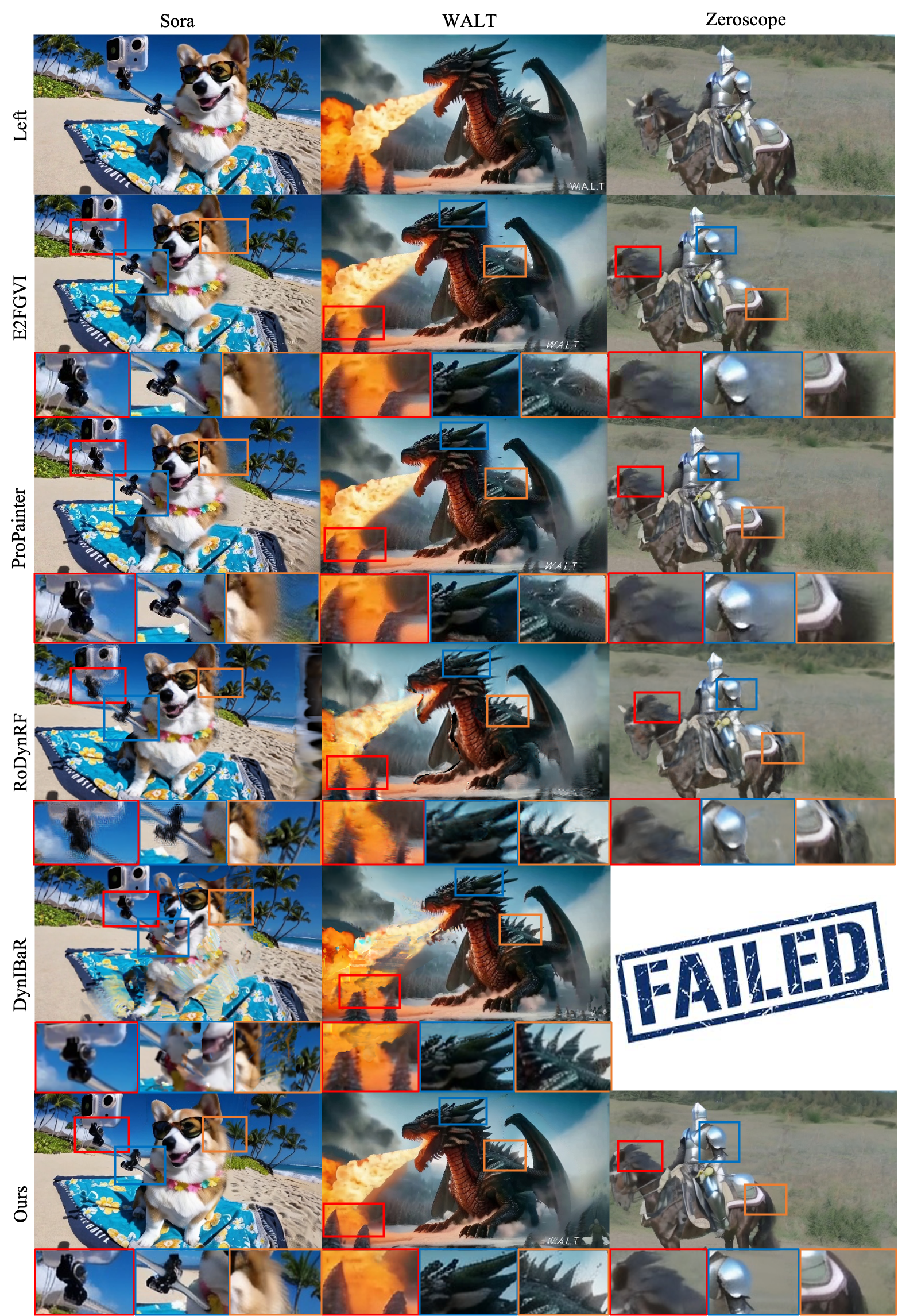}
    \caption{\textbf{Qualitative comparisons.} The first row shows left-view images. The video inpainting methods E2FGVI and ProPainter tend to generate blurry content in disoccluded regions, such as knight's arm and corgi's face. RoDynRF lacks the generation ability, thus content on the right side of the corgi case is poor. DynIBaR's results contain artifacts, and it requires camera poses as inputs, which failed in some scenarios. On the contrary, our method takes advantages of video generation models and is pose-free, thus generates high-quality content in different scenarios.}
    \label{fig:comparisons}
\end{figure}


\subsection{Qualitative Results}
\textbf{Qualitative Comparisons.} We show qualitative comparisons in Fig.~\ref{fig:comparisons}. Previous video inpainting methods suffer from a common problem -- the generated content in disoccluded regions is blurry, such as the knight's arm, horse's tail, and corgi's face, presumably because that these methods are trained on limited datasets.
On the other hand, novel view synthesis methods suffer from unstable camera pose estimation (\textit{e.g.}, DynIBaR fails on some videos).
Though good at reconstructing visible content from the monocular video, they are typically poor at synthesizing novel contents in the disoccluded regions that are not observed in any frames (\textit{e.g.},  ghost effect near the boundary in the RoDynRF result on the corgi example).
In contrast, our approach takes advantage of generative capability of video diffusion models trained on massive scale datasets and does not require camera poses of the input video as inputs, thereby generating high-quality content in various types of scenarios (last row of Fig.~\ref{fig:comparisons}) and consistently outperforms baseline methods.
Additionally, we visualize the stereo effects of different methods on the corgi case using a stereo depth estimator~\cite{Li_2021_ICCV}, which predicts disparity values from the stereo images. As shown in Fig.~\ref{fig:3D_effects}, RoDynRF and DynIBaR exhibit less depth variation, indicating weaker stereo effects. This occurs when the camera is wrong and training process overfits the training views, resulting in a sub-optimal 3D representation. 

\subsection{Quantitative Results}
In this part, we show quantitative comparisons with other baselines.
We primarily rely on a dedicatedly designed user study to evaluate the quality of generated stereoscopic video on various quality axes.
We also provide an objective metric to measure the semantic similarity between the left and right views using pre-trained CLIP models.


\textbf{Human Perception.}
To assess the perceived visual quality, we conducted a user study with 20 participants (9 female, age $\mu=33, \sigma=6.2)$. On a VR headset, each participant viewed and evaluated five generated videos (out of 20 in total) by all five methods on stereo effect, temporal consistency, image quality, and overall experience using a 7-point Likert scale \cite{likert1932technique}. A total of 435 evaluations (DynIBaR failed to generate 13 videos) were counterbalanced and randomly shuffled. We also included a training session to eliminate novelty effects. Results are summarized in Table~\ref{tab:quantitative}, with details in the supplemental material. Our method outperforms other baselines on measured metrics.

\begin{table}[!htb]
    \centering
    \begin{tabular}{cccccc}
    \toprule
    & E2FGVI & ProPainter & RoDynRF & DynIBaR & Ours\\
    \midrule
    Stereo Effect~$\uparrow$  & 4.79 (1.08) & 4.81 (1.13) & 2.97 (1.34) & 1.86 (1.25) & \textbf{5.24} (0.94) \\
    Temporal Consistency~$\uparrow$  & 4.74 (1.33) & 4.74 (1.22) & 3.35 (1.66) & 1.89 (1.33) & \textbf{5.15} (1.22) \\
    Image Quality~$\uparrow$  & 4.42 (1.27) & 4.38 (1.28) & 2.84 (1.60) & 1.67 (1.07) &  \textbf{5.12} (1.33) \\
    Overall Experience~$\uparrow$  & 4.67 (1.04) & 4.66 (1.09) & 2.92 (1.43) & 1.72 (1.06) &  \textbf{5.35} (0.99)\\
    \bottomrule 
    \end{tabular}
    \caption{\textbf{Quantitative comparisons.} This table reports results of human perception experiments as mean (std). Our method outperforms other baselines on all metrics. Kruskal-Wallis tests~\cite{kruskal1952use} reveal significant effects of group on all metrics ($\chi^2 > 13.3, p < 0.001 ^{***}$). Post-hoc tests using Mann-Whitney tests~\cite{mann1947test} with Bonferroni correction reveal significant effects ($p < 0.05 ^{*}, |r| > 0.1$) for each pairwise comparison, except E2FGVI \textit{vs.} ProPainters yield comparable results. }
    \vspace{-10pt}
    \label{tab:quantitative}
\end{table}

\textbf{Semantic Consistency.} 
We additionally check the semantic consistency between the left and the right view.
We use pre-trained CLIP model~\cite{radford2021learning} to extract features for both left views and right views of a stereoscopic video, and then calculate the feature distance following Sun \emph{et al.}~\cite{taited2023CLIPScore} to obtain the semantic consistency score. 
In Table~\ref{tab:ablation_study}, our method attains the best semantic consistency (96.44) over other baselines.


\vspace{-5pt}
\subsection{Ablation Studies}
\begin{figure}[!htb]
    \centering
    \includegraphics[width=1.0\linewidth]{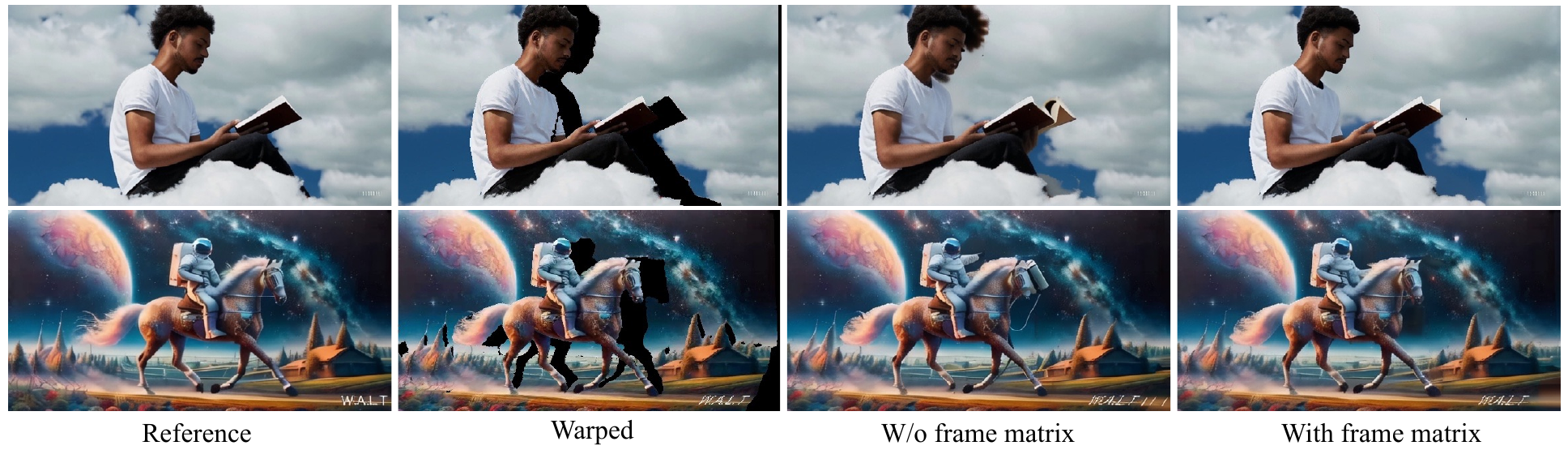}
    \vspace{-15pt}
    \caption{\textbf{Semantically consistent content generation.} The reference frames are warped into the target view with disoccluded regions set to be black. Without using frame matrix, the generated content does not match the reference, such as the book and the face of horse. With frame matrix, the inpainted contents are more semantically reasonable.}
    \label{fig:sematic_matching}
    \vspace{-10pt}
\end{figure}

\textbf{Effect of Frame Matrix.} In Fig.~\ref{fig:sematic_matching}, we showcase that using frame matrix benefits semantic consistency between the left and right views. 
Without using frame matrix, the disoccluded regions in warped images can be inpainted with unconstrained contents, which are likely to be inconsistent with the left view given impressive generative capability of the diffusion model, such as the hair of the man and the head of the horse. This is also revealed in Table~\ref{tab:ablation_study}, where CLIP Score drops from 96.44 to 95.81 when disabling frame matrix. Thanks to constraints from other frames within the frame matrix, our method generates both reasonable foreground and background contents in the disoccluded regions.
More studies of frame matrix are included in Sec.\ref{sec:more_frame_matrix} of the supplemental material. 

\begin{figure}[!htb]
    \centering
    \includegraphics[width=1.0\linewidth]{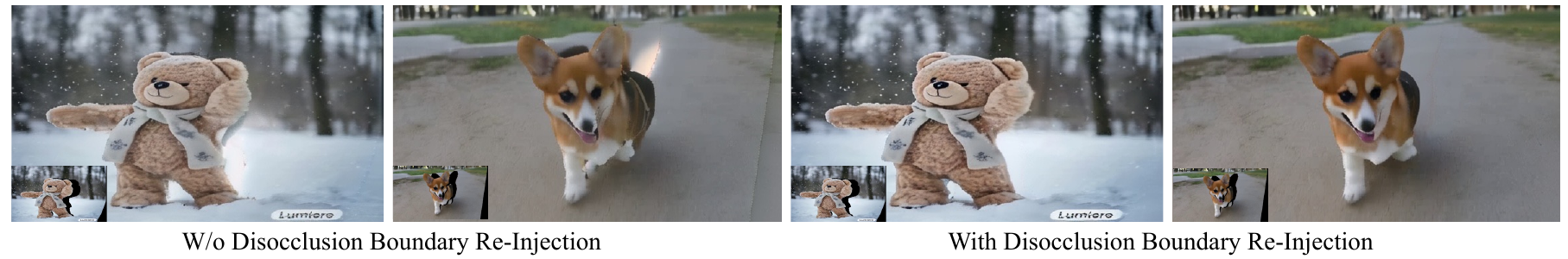}
    \vspace{-15pt}
    \caption{\textbf{Disocclusion Boundary Re-injection.} Without disocclusion boundary re-injection, the inpainted images usually contain artifacts. Bottom-left corner shows the warped image.}
    \vspace{-10pt}
    \label{fig:update_feature}
\end{figure}

\paragraph{Effects of Disocclusion Boundary Re-Injection.}
In Fig.~\ref{fig:update_feature}, we demonstrate the importance of updating unoccluded latent features for high-quality results. Without this update, the disoccluded region is inpainted with unnatural textures that don't blend well with the surrounding content. With the update, the inpainted content blends seamlessly. This is reflected quantitatively in Table~\ref{tab:ablation_study}, where the CLIP Score drops from 96.44 to 95.60 when unoccluded feature updates are discarded.

\vspace{-5pt}

\vspace{-5pt}
\begin{table}[!htb]
    \centering
    \begin{tabular}{cccccccc}
    \toprule
    Method & E2FGVI & ProPainter & RoDynRF & DynIBaR & Ours - FM & Ours - DBR & Ours\\
    \midrule
    CLIP~$\uparrow$ & 94.34 & 95.29 & 96.03 & 93.24 & 95.81 & 95.60 & \textbf{96.44}\\
    \bottomrule
    \end{tabular}
    \caption{\textbf{Semantic consistency score}. We show the semantic consistency using CLIP feature similarity~\cite{hessel2021clipscore} between the left and right view. Our method outperforms previous methods as well as ablated cases.}
    \label{tab:ablation_study}
    \vspace{-15pt}
\end{table}

\paragraph{Different Stereo Baselines.}

\begin{figure}[!htb]
    \centering
    \includegraphics[width=1.0\linewidth]{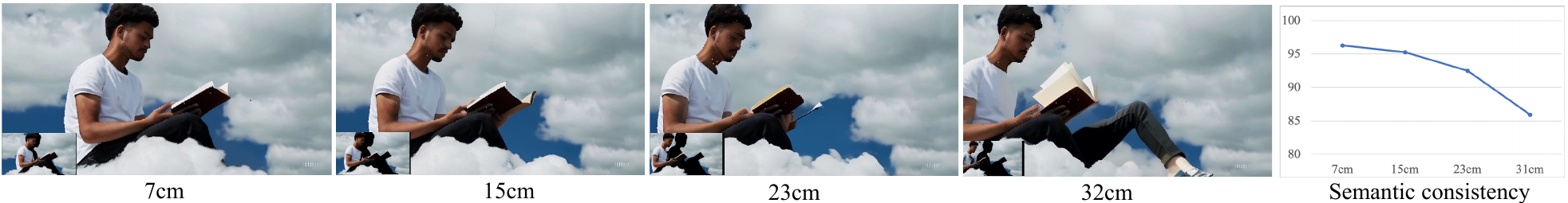}
    \vspace{-15pt}
    \caption{\textbf{Result with different stereo baselines.}
Unnatural artifacts begin to appear as the baseline expands. Our method performs well for stereoscopic video generation where baseline is usually less than 7cm.}
\vspace{-10pt}
    \label{fig:stereo_baslie}
\end{figure}

Fig.~\ref{fig:stereo_baslie} shows increasing the stereo baseline makes inpainting harder and degrades stereoscopic video quality, as reflected by CLIP score. Our method is resilient to larger baselines, failing beyond ~20cm (depth normalized to 1.0-10.0m). This range is sufficient for generating 3D stereoscopic video for most people, given typical inter-pupillary distances of 5-7cm.

\section{Limitations}
Although our results demonstrate the possibility of generating 3D stereoscopic videos using pre-trained video diffusion models, challenges remain. For one, we did not study longer videos because the architecture of a typical video diffusion model supports generating videos only a couple of seconds long. One possible solution for long 3D stereoscopic video generation is to use stronger foundational models, such as Sora~\cite{videoworldsimulators2024}. Alternatively, we could gradually generate longer videos by overlapping frames of shorter videos. Additionally, our method is dependent on a depth estimation model~\cite{yang2024depth}, which may fail, \textit{e.g.}, when dealing with thin structures.

\section{Conclusion}
We proposed a complete system for stereoscopic video generation, using a video diffusion model and our \textit{frame matrix} inpainting scheme. Given the fast adoption of video generation, our approach bridges the gap between current ability to generate monocular and stereoscopic videos. In particular, we showed that our \textit{frame matrix} formulation significantly advances the state-of-the-art for generative stereoscopic video, and can be adopted by existing and future video diffusion models.

\small{
\bibliographystyle{plain}
\bibliography{egbib}
}

\clearpage
\newpage
\appendix
\section*{Appendix}

In the supplementary sections, we provide more studies and details of the proposed method. 
\begin{itemize}
    \item In sec.~\ref{sec:pseudocode}, we provide a pseudocode describing our frame matrix inpainting. 
    \item In sec.~\ref{sec:details_data}, it includes details of data preprocessing in handling warping-related artifacts. 
    \item In sec.~\ref{sec:details_user_study}, we include details of human perception experiments and provide additional comparisons with Deep3D.
    \item In sec.~\ref{sec:more_frame_matrix}, it contains more studies of frame matrix, including different trajectories in frame matrix and consistency across different views.
    \item In sec.~\ref{sec:more_results}, we display more results in different scenarios.
    \item In sec.~\ref{sec:more_ablation}, we show the effectiveness of our data preprocessing. 
\end{itemize}
More video results and comparisons can be found in the supplementary webpage (.html).

\section{Algorithm Details}
\label{sec:pseudocode}
In the algorithm below, we present the detailed steps to denoise the Frame Matrix with spatial-temporal resampling, where we set $\mu_{\theta}(\VidLatent_{t}, \Prompt, t) = \frac{1}{\sqrt{1-\beta_t}}(\VidLatent_t - \frac{\beta_t}{\sqrt{1-\Bar{\alpha}_t}}\epsilon_{\theta}(\VidLatent_t,\Prompt,t))$, following DDPM\cite{ho2020denoising}.

\begin{algorithm}
  \caption{Frame Matrix Inpainting}
  \label{algo_fm}
  \begin{algorithmic}
    \STATE \textbf{Input: } $\VidLatent_T \sim~ \mathcal{N}(\mathbf{0}, \mathbf{I})$: Initial noisy latent maps \\
    $\VidLatent_0$: Initial clean latent maps \\
    \FOR{$t=T,...,1$}
        \FOR{$n=1,...,N$}
            \IF{n is odd}
                \STATE Denoise time sequences $\{\VidLatent_{(\FrameSub,:)t} | \FrameSub = 1,..., \FrameSubMax \}$:
                \FOR{$\FrameSub = 0,..,\FrameSubMax$}
                   \STATE $\VidLatent_{(\FrameSub,:)t-1}^{\text{known}} \sim \mathcal{N}(\sqrt{\NoiseScale}\VidLatent_{(\FrameSub,:)0}, (1 - \NoiseScale)\mathbf{I})$ \\
                   \STATE $\VidLatent_{(\FrameSub,:)t-1}^{\text{unknown}} \sim \mathcal{N}(\mu_{\theta}(\VidLatent_{(\FrameSub,:)t}, \Prompt, t), \Sigma_{\theta}(\VidLatent_{(\FrameSub,:)t}, \Prompt, t))$ \\
                    \STATE $\VidLatent_{(\FrameSub,:)t-1} = \LatentMasks_{(\FrameSub,:)} ~\odot~ \VidLatent_{(\FrameSub,:)t-1}^{\text{known}} + (1-\LatentMask_{(\FrameSub,:)}) ~\odot~ \VidLatent_{(\FrameSub,:)t-1}^{\text{unknown}}$
                \ENDFOR
            \ELSE
                \STATE Denoise view sequences
                $\{\VidLatent_{(:,\ViewSub)t} | \ViewSub = 1,..., \ViewSubMax\}$:
                \FOR{$\ViewSub = 0,..,\ViewSubMax$}
                   \STATE $\VidLatent_{(:,\ViewSub)t-1}^{\text{known}} \sim \mathcal{N}(\sqrt{\NoiseScale}\VidLatent_{(:,\ViewSub)0}, (1 - \NoiseScale)\mathbf{I})$ \\
                   \STATE $\VidLatent_{(:,\ViewSub)t-1}^{\text{unknown}} \sim \mathcal{N}(\mu_{\theta}(\VidLatent_{(:,\ViewSub)t}, \Prompt, t), \Sigma_{\theta}(\VidLatent_{(:,\ViewSub)t}, \Prompt, t))$ \\
                    \STATE $\VidLatent_{(:,\ViewSub)t-1} = \LatentMasks_{(:,\ViewSub)} ~\odot~ \VidLatent_{(:,\ViewSub)t-1}^{\text{known}} + (1-\LatentMask_{(:,\ViewSub)}) ~\odot~ \VidLatent_{(:,\ViewSub)t-1}^{\text{unknown}}$
                \ENDFOR
            \ENDIF
             \STATE Add back one noise step for resampling:\\
                        \STATE $\VidLatent_{t} \sim \mathcal{N}(\sqrt{1-\beta_{t-1}}\VidLatent_{t-1}, \beta_{t-1}\mathbf{I})$
        \ENDFOR
    \ENDFOR
  \end{algorithmic}
\end{algorithm}

\vspace{-5pt}
\begin{figure}[!htb]
    \centering
    \includegraphics[width=1.0\linewidth]{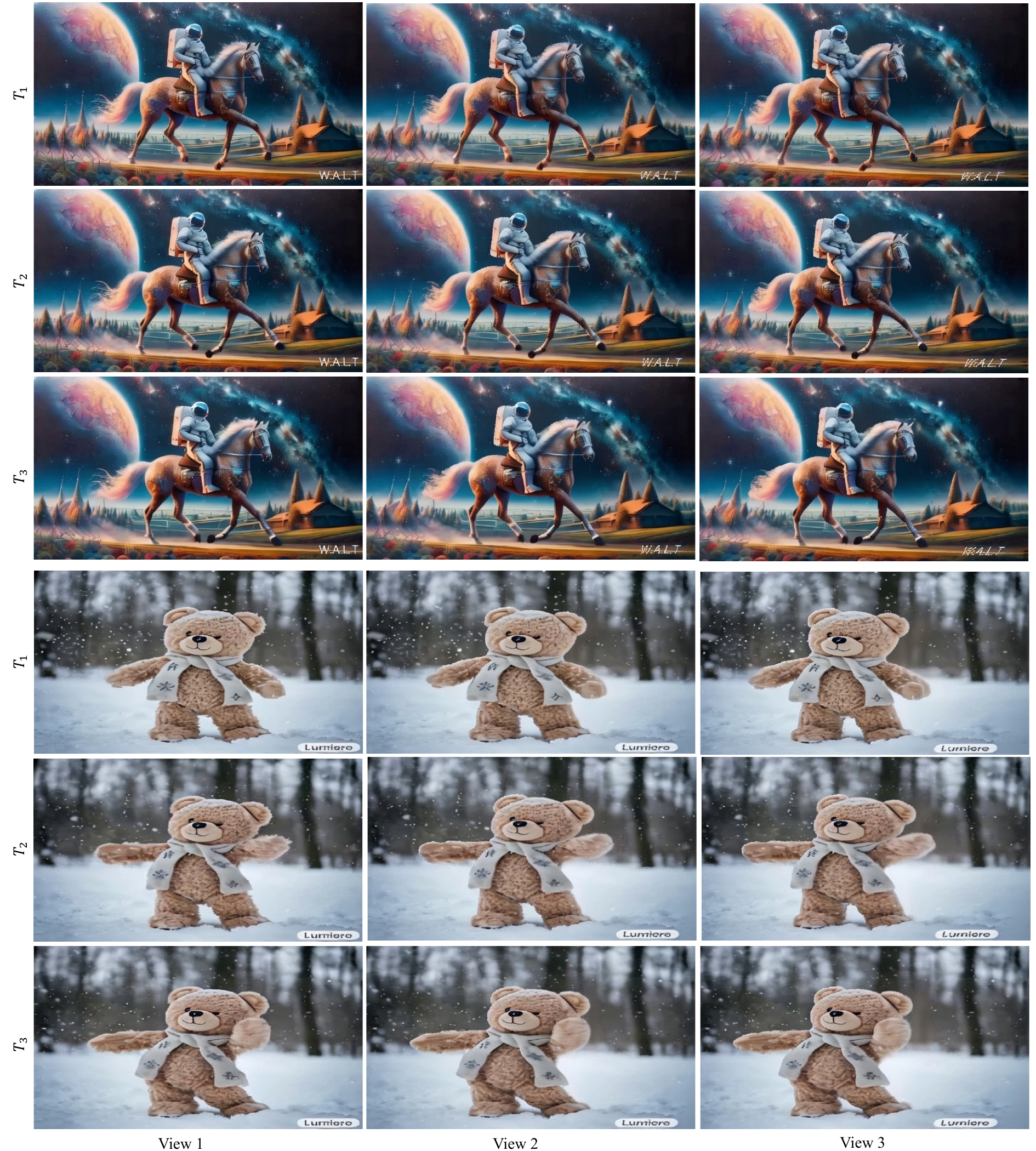}
    \vspace{-10pt}
    \caption{\textbf{Videos in frame matrix.} In both cases, each column is a generated video in a camera, and each row represents generated frames in different cameras at a specific timestamp.}
    \vspace{-10pt}
    \label{fig:multi_view}
\end{figure}

\vspace{-5pt}
\begin{figure}[!htb]
    \centering
    \includegraphics[width=1.0\linewidth]{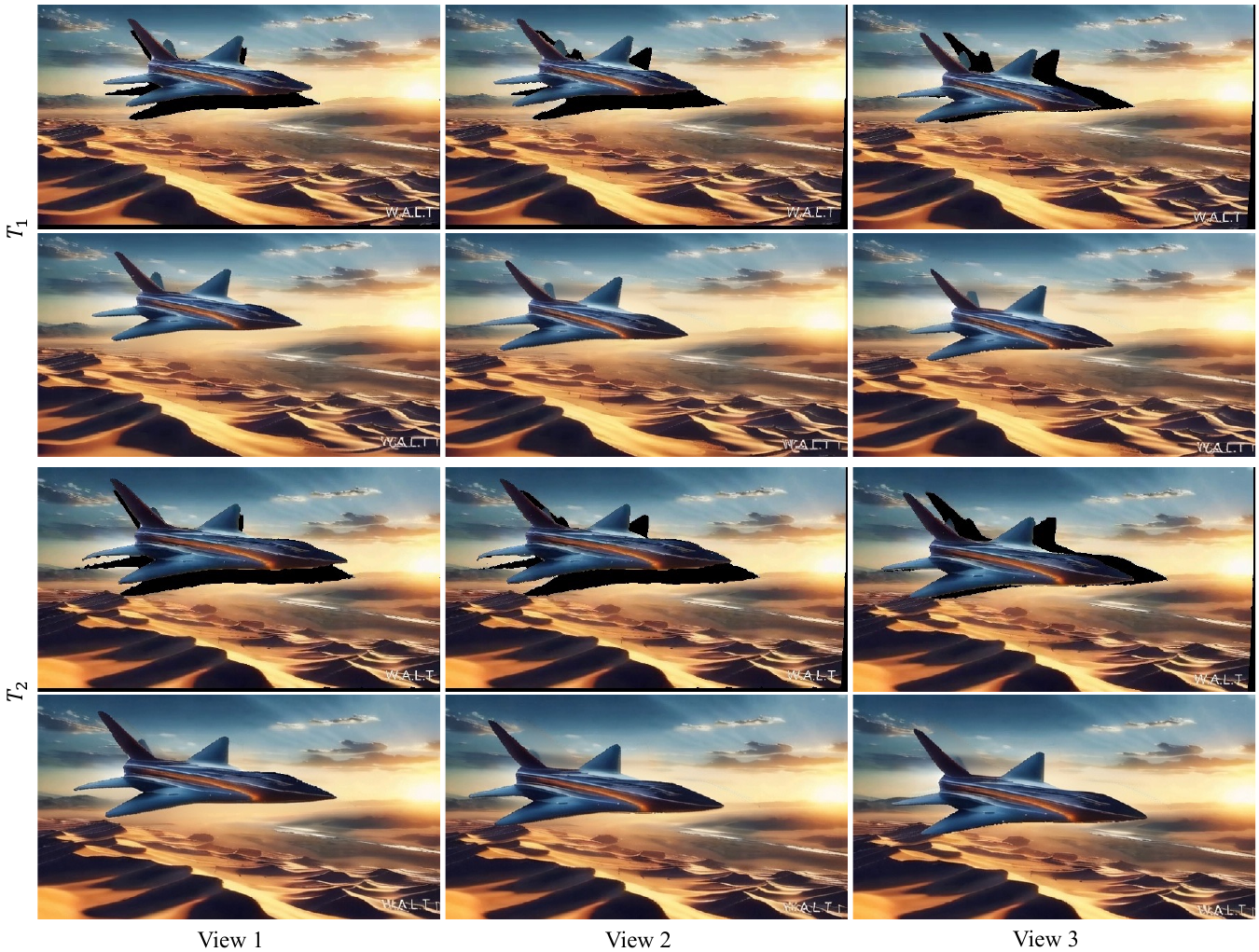}
    \vspace{-10pt}
    \caption{\textbf{Videos in frame matrix constructed using a spiral trajectory.} Warped and generated frames in different cameras at different timestamps.}
    \vspace{-5pt}
    \label{fig:multi_view_traj}
\end{figure}

\begin{figure}[!htb]
    \centering
    \includegraphics[width=1.0\linewidth]{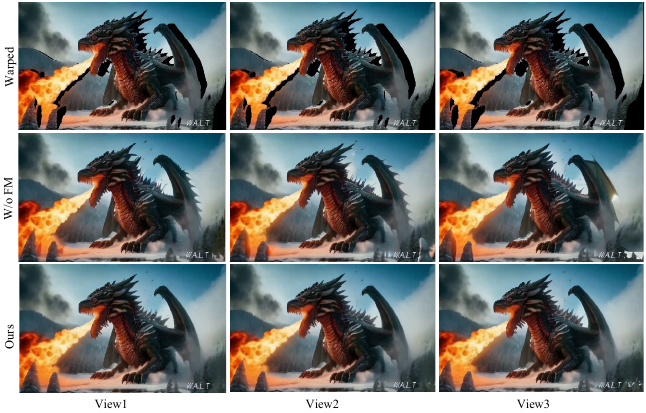}
    \vspace{-10pt}
    \caption{\textbf{Consistency.} The content is inconsistent when each view is generated independently. Frame matrix benefits the consistency of our results across different views. Please note the dragon's wing.}
    \vspace{-10pt}
    \label{fig:consistency}
\end{figure}

\begin{figure}[!htb]
    \centering
    \includegraphics[width=1.0\linewidth]{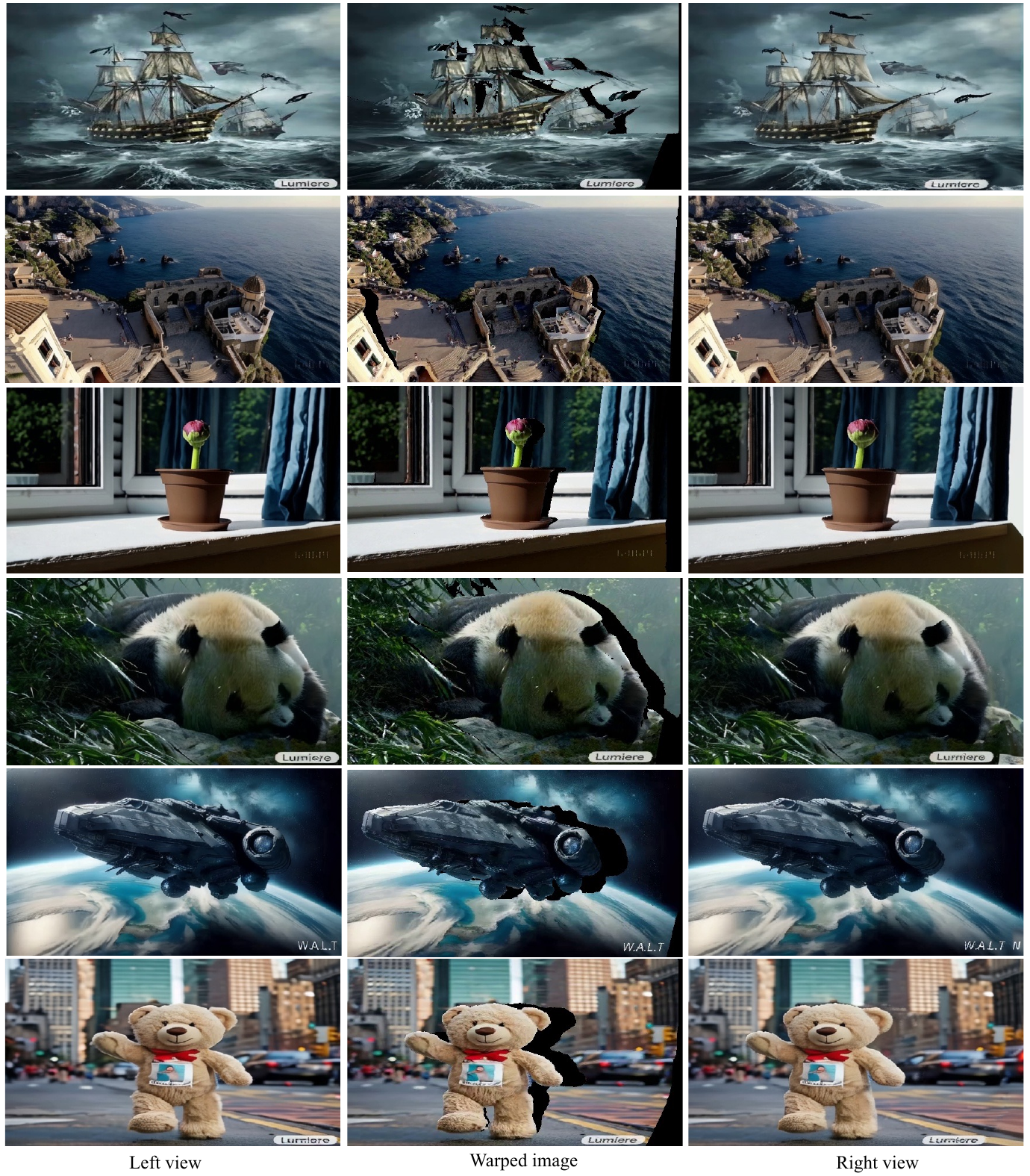}
    \vspace{-10pt}
    \caption{\textbf{More results.} We display more generated results in different scenarios.}
    \vspace{-5pt}
    \label{fig:more_results}
\end{figure}

\begin{figure}[!htb]
    \centering
    \includegraphics[width=1.0\linewidth]{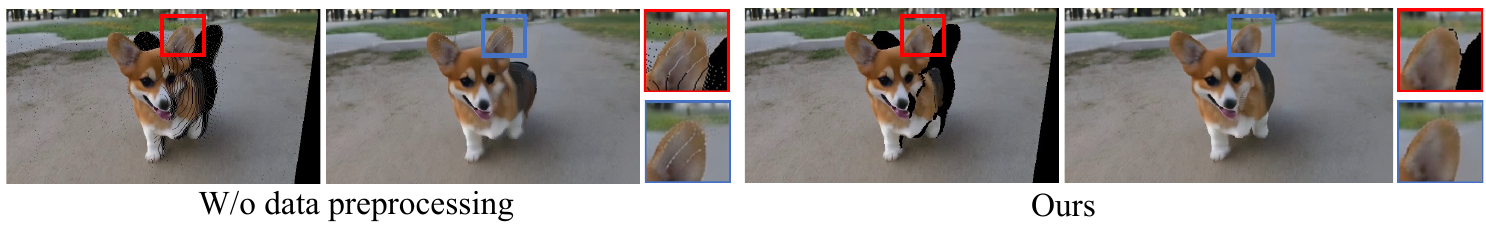}
    \vspace{-10pt}
    \caption{\textbf{Data preprocessing.} Left: without handling isolated points and entangled foreground and background (the gray road can be seen through the dog's ear) in warped images, these artifacts remain in the final results. Right: our results have no artifacts.}
    \vspace{-10pt}
    \label{fig:ab_data_preprocessing}
\end{figure}

\vspace{-5pt}
\begin{figure}[!htb]
    \centering
    \includegraphics[width=1.0\linewidth]{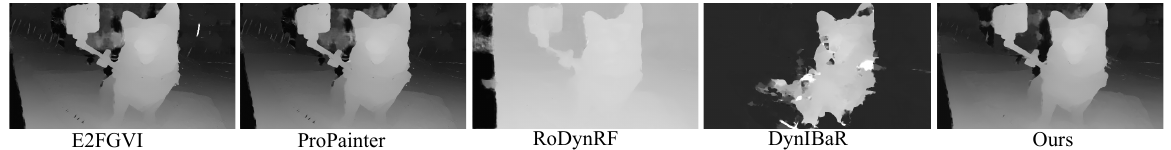}
    \vspace{-15pt}
    \caption{\textbf{Disparities.} We visualize stereo effects by predicting disparity values from stereo images~\cite{Li_2021_ICCV}.}
    \label{fig:3D_effects}
    \vspace{-10pt}
\end{figure}

\section{Details of Data Preprocessing}
\label{sec:details_data}
\textbf{Multi-Plane projection.} 
Given RGB-D images, we warp them into a target camera view. Instead of projecting all pixels onto one image plane and handling occlusions using z-buffer, we divide the camera view space into multi-plane images ${\{I_1^{step0}, ..., I_N^{step0}\}}$ (N=4 in this paper) according to near and far depths, then each pixel is projected onto the image plane closest to it. We use ${\{M_1^{step0}, ..., M_N^{step0}\}}$ to indicate valid pixel positions on each image plane. By doing this, the foreground and background are separated in different planes temporarily, which makes dealing with artifacts (i.e., isolated points and entangled foreground and background content in Fig.~\ref{fig:ab_data_preprocessing} left) easier. 

\textbf{Remove isolated points.} Due to the inaccuracy of depth values around image boundaries, these pixels are warped into wrong positions leading to isolated pixels (see red box in Fig.~\ref{fig:ab_data_preprocessing} left). Intuitively, isolated pixels have no or very few neighbors, thus we detect isolated pixels based on this observation. Specifically, we apply convolution on each mask plane $M_i^{step0}$ using a $3\times3$ kernel, after which isolated pixels are empirically determined where values after convolution are less than 0.5. We remove these isolated pixels on both RGB and mask planes to obtain new ${\{I_1^{step1}, ..., I_N^{step1}\}}$ and ${\{M_1^{step1}, ..., M_N^{step1}\}}$.

\textbf{Handle foreground and background entanglement.} Since the depth image is not a watertight representation, the warped image usually contains small cracks/holes that confuse foreground and background content. For example, the gray road can be seen through the dog's ear in Fig.~\ref{fig:ab_data_preprocessing} left. Similar to handling isolated pixels, we use a $3\times3$ Gaussian kernel to perform convolution on each mask plane $M_i^{step1}$. When there are cracks, the values after convolution will be less than 1. In this paper, positions with no pixel values (0 in $M_i$) but with greater values than 0.2 after convolution are considered cracks. We fill these cracks via interpolating nearby valid pixels in each image plane and obtain new multi-plane images ${\{I_1^{step2}, ..., I_N^{step2}\}}$ and ${\{M_1^{step2}, ..., M_N^{step2}\}}$.

After handling artifacts in each image plane, all image planes are blended into one image (e.g., Fig.~\ref{fig:ab_data_preprocessing} ours left) in a back-to-front order using Eq.~\ref{eq:blend}, where the content of front plane blocks content belongings to the plane at the back. 

\begin{equation}
    I = I\times(1-M_i^{step2}) + I_i^{step2}\times M_i^{step2},\ for\ i\ in\ [N, ..., 1].
    \label{eq:blend}
\end{equation}


\vspace{-5pt}
\begin{figure}
    \centering
    \includegraphics[width=1.0\linewidth]{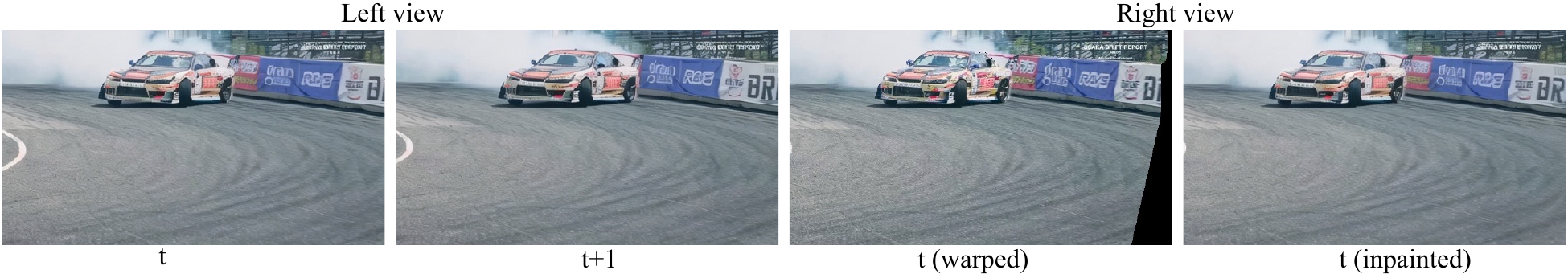}
    \vspace{-15pt}
    \caption{\textbf{Ability to utilize unobserved content.} Left view: two consecutive images observed by the left view. Right view: the warped and inpainted images at time t. Note that the black region is inpainted with the character ``R'', matching the characters in the second image at time t+1.}
    \label{fig:borrow}
    \vspace{-5pt}
\end{figure}

\vspace{-5pt}
\begin{figure}
    \centering
    \includegraphics[width=1.0\linewidth]{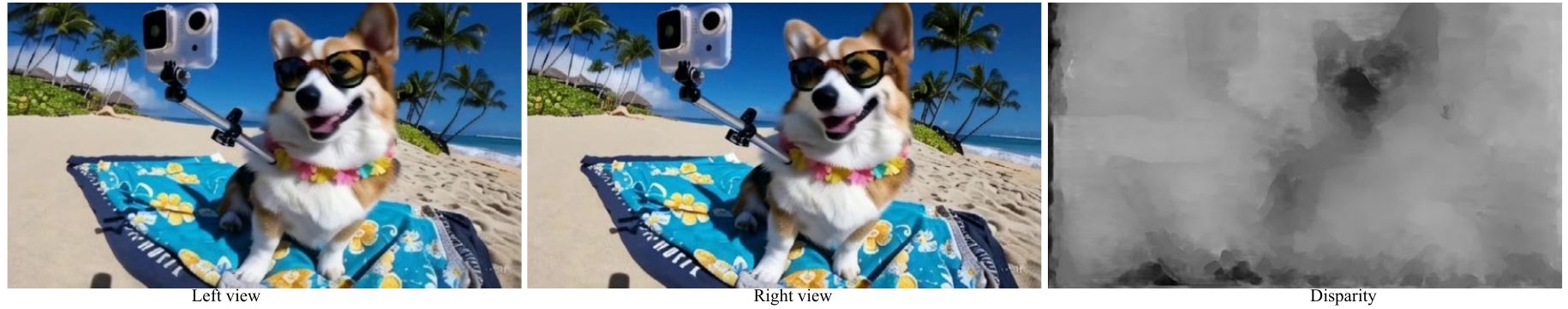}
    \vspace{-15pt}
    \caption{\textbf{Results of Deep3D.} Deep3D does not provide the function to change the stereo baseline, and the vague disparity map on the right side demonstrates its weak stereo effects.}
    \label{fig:deep3d}
    \vspace{-5pt}
\end{figure}

\section{Details of Human Perception Study}
\label{sec:details_user_study}

\textbf{Participants.}
To evaluate the perceived quality of the generated stereoscopic videos, we recruited 20 participants (9 females) at least 18 years old ($\mu=33, \sigma=6.2$) with normal or corrected-to-normal vision at an anonymous institution via email lists and group communication software. The majority of participants had some experience with virtual reality.
None of the participants was involved with this project prior to the user
study.

\textbf{Study setup.}
The study was conducted in a quiet meeting room with a commercial VR headset as the primary apparatus. The study software is implemented in Unity 2023.3.0b and we render stereoscopic videos with custom shaders on a $1.8m \times 1.0m$ quad that is three meters away from the participant in the world space, which occupies approximately 33.4 degrees in width and 18.92 degrees in height initially. Users have the freedom to move themselves within the meeting room to examine the stereoscopic video. This setup allowed participants to experience the stereoscopic videos in virtual reality settings and provided a controlled environment for the user study. 

\textbf{Study protocol.}
Each study session consists of a demographics interview with consent forms, a training session, and an evaluation session. To eliminate the ordering effect, 
we randomly counterbalanced all five methods for each video and assigned five random videos (out of 20 videos) with five conditions to each participant. However, since DynIBaR method failed to generate 13 videos, we collected a total of $5 \times 5 \times 20 - 13 \times 5 = 435$ evaluations from 20 participants, resulting in $100$ human evaluations for each method except DynIBaR. During the training session, we randomly picked a video that was outside of the assigned videos to the participant and asked the participant to rate the stereoscopic effect, temporal consistency, graphical quality, and overall experience on a 7-point Likert scale \cite{likert1932technique}, with 1 being the lowest, 7 being the highest, and 4 being the average. This procedure helps eliminate the novelty effect and calibrate the user's rating before the formal evaluation session.
In the formal evaluation, we prompted the participant with the question like \textit{``How would you like to rate the stereoscopic effect of the video on a 7-point scale, with 1 being the lowest, 7 being the highest, and 4 being the average?''} and asked the user the reason behind the rating.

\textbf{Metrics.}. We evaluate the perceived quality of generated stereo videos based on three key aspects: 1. Stereo Effect. This refers to the perception of depth achieved by presenting slightly different images to each eye. A strong stereo effect makes objects appear closer or farther away, enhancing the 3D experience. Example questions: "How strong was the 3D effect in the video?" and "Which video felt more immersive due to the 3D effect?" 2. Temporal Consistency. This aspect assesses the smoothness of scene motion and the absence of artifacts such as jitter or ghosting over time. Example questions: "How smooth and natural did the motion of objects appear?" and "Did you notice any flickering, jumpiness, or distortions in the video?" 3. Graphical Quality. This evaluates the overall visual appeal of the video, including the quality of details, textures, lighting, and color fidelity. Example questions: "How would you rate the visual quality of the video?" and "Which video had more detailed and realistic textures?"

\textbf{Study results.}
Overall, despite the missing data points for the DynIBaR method in some videos, Kruskal-Wallis tests~\cite{kruskal1952use} reveals significant effects of group on all metrics respectively ($\chi^2 > 13.3, p < 0.01$): with stereo effect $\chi^2 = 186.3, p < 0.001$, temporal consistency $\chi^2 = 121.3, p < 0.001$, graphical quality $\chi^2 = 153.2, p < 0.001$, and overall experience $\chi^2 = 192.9, p < 0.001$. We further performed post-hoc tests using Mann-Whitney tests~\cite{mann1947test} with Bonferroni correction, which revealed significant effects ($p < 0.05, |r| > 0.1$) for each pairwise comparison, except E2FGVI \textit{vs.} ProPainters. Specifically, for Ours \textit{vs.} E2FGVI, $p=0.002$ on stereo effect, $p=0.030$ on temporal consistency, $p<0.001$ on graphical quality and overall experience. For Ours \textit{vs.} ProPainter, $p=0.004$ on stereo effect, $p=0.017$ on temporal consistency, $p<0.001$ on graphical quality and overall experience.


\textbf{Study findings.} Our results suggest that our methods achieve significantly better perceived stereoscopic effect than all other methods, while improvement in graphical quality and overall experience is more evident over stereoscopic effect; and stereo effect more evident over temporal consistency. During the study, we also observed many positive comments about our methods like ``the contour is more clear'', \textit{``the graphics are sharper with fewer artifacts''}; however, we also observed negative or neutral feedback like \textit{``some part really works and some parts don't: one side of the turtle face is wrong''}, and \textit{``I see no difference (on the faces)''} from two participants. This suggests future research to investigate holistic perceptual consistency in stereoscopic videos and finetune models for special subjects like human beings.

\textbf{Additional User Study on Ours \textit{vs.} Deep3D.}

Despite that we did not include Deep3D in the design of our initial user study, we further conducted a human evaluation between Ours and Deep3D across the same metrics with a total of 190 random evaluations over 20 random videos, following the same protocol. Pairwise Mann-Whitney tests with Bonferroni correction reveal significant effects on stereo effect ($p < 0.001$), overall experience ($p < 0.001$), and temporal consistency $(p = 0.015)$. We found our method outperforms Deep3D in stereo effect and overall experience, yet falling slightly short in temporal consistency.

Similar to Fig.4 in main paper, we visualize Deep3D's disparity map in Fig.~\ref{fig:deep3d}. The vague disparity map in the third column demonstrates weak stereo effects, which matches the statistic results in Table~\ref{tab:quantitative}. By manually modifying the disparity map or changing the stereo baseline, 3D effects may become apparent. However, Deep3D does not support these functions. 

\begin{table}[!htb]
    \centering
    \begin{tabular}{cccccc}
    \toprule
    & Deep3D & Ours & $p$-value & $|r|$ (effect size)\\
    \midrule
    Stereo Effect~$\uparrow$  & 2.29 (1.63) & \textbf{5.29} (1.09) & < 0.001 $^{***}$ & 0.60 (large) \\
    Temporal Consistency~$\uparrow$  & \textbf{5.37} (1.23) & 5.06 (1.25) & 0.015 $^*$ & 0.49 (medium)\\
    Image Quality~$\uparrow$  & \textbf{5.27} (1.17) & 5.12 (1.19) & 0.103 & 0.10 (small)\\
    Overall Experience~$\uparrow$  & 3.68 (1.36) & \textbf{5.08} (1.09) & < 0.001 $^{***}$ & 0.57 (large)\\
    \bottomrule 
    \end{tabular}
    \caption{\textbf{Quantitative comparisons.} This table reports results of human perception experiments as mean (std)  between Deeph3D and Ours. Our method outperforms Deeph3D in stereo effect and overall experience, yet falls slightly short in temporal consistency. Mann-Whitney tests with Bonferroni correction reveals significant effects on stereo effect ($p < 0.001$, $Z=-8.24$), overall experience ($p < 0.001$, $Z=-7.92$), and temporal consistency $(p = 0.015, Z=-6.72)$. }
    \label{tab:quantitative}
\end{table}

\section{More Results of Frame Matrix}
\label{sec:more_frame_matrix}
\textbf{Other trajectories in frame matrix.} In main paper, we show generated 3D left and right views. Here, we additionally show the results of other trajectories. In Fig.~\ref{fig:multi_view}, we selectively display frames generated within the frame matrix at different timestamps (3 out of 16) in different camera views (3 out of 8). From the results, both foreground and background content are coherent across different frames. Moreover, instead of constructing frame matrix using cameras moving from left to right, we alternatively move the camera following a spiral trajectory. In Fig.~\ref{fig:multi_view_traj} first and third rows, we selectively show the warped images in different camera views (3 out of 16), where disocclusions appear around the plane. Under each warped image, we display the corresponding image with disocclusions filled. 


\textbf{Consistency.} In Fig.~\ref{fig:consistency}, the first row is warped images under different camera views. We generate each view independently and show results in the second row, where the content is not consistent across different views, such as the dragon's wing. With the help of the frame matrix, which also regularizes generation in the direction of camera motion, our results in the third row are more consistent.

\section{More Results of Stereoscopic Videos}
\label{sec:more_results}
\textbf{More cases.} In this part, more generated results are displayed in Fig.~\ref{fig:more_results}. The proposed method works on different scenarios, such as the beautiful church, imaginary scenes, and ships in the storm where the whole scene is dynamic. The high-quality generated results in Fig.~\ref{fig:more_results} right column demonstrate the generalization ability of the proposed method. 

\textbf{Ability to utilize temporal context for inpainting.}
Our method is able to harmonize image contents between different temporal frames during inpainting and thus enhance temporal consistency.
Figure~\ref{fig:borrow} shows one example.
When inpainting the right-view frame at $t$, our method successfully creates content that is consistent with the left-view frame at $t+1$ (see the generated character ``R'' in the disoccluded region).
Note that such consistency is maintained automatically thanks to frame matrix based denoising, since all temporal frames are taken into account.


\section{More Ablation Studies}
\label{sec:more_ablation}
\textbf{Effects of Data Preprocessing.}
In Fig.~\ref{fig:ab_data_preprocessing} left, obvious artifacts are in warped images, such as isolated points and cracks where the foreground ear is entangled with the background gray road, and these artifacts remain in the final generated results. On the contrary, Fig.~\ref{fig:ab_data_preprocessing} right shows our results, which are artifacts-free after applying the proposed data preprocessing.

\end{document}